\pdfoutput=1

\documentclass[11pt]{article}

\usepackage[]{acl}

\usepackage{times}
\usepackage{latexsym}

\usepackage[T1]{fontenc}
\usepackage[utf8]{inputenc}
\usepackage{microtype}
\usepackage{inconsolata}
\usepackage{booktabs}
\usepackage{multirow}
\usepackage{enumitem}
\usepackage{xspace}
\usepackage{amsfonts}
\usepackage{amsmath}
\usepackage[subrefformat=parens,labelformat=parens]{subfig}
\usepackage{graphicx}
\usepackage{amsthm}
\usepackage{algorithm}
\usepackage[noend]{algpseudocode}

\newcommand{\vc}[2]{#2{(#1)}}
\DeclareMathOperator*{\argmax}{arg\,max \quad}

\newcommand{\allminilm}{\textsc{AllMiniLM}\xspace}
\newcommand{\gtebase}{\textsc{Gte-Base}\xspace}
\newcommand{\gtelarge}{\textsc{Gte-Large}\xspace}
\newcommand{\ccsmall}{\textsc{CC3M}\xspace}
\newcommand{\cclarge}{\textsc{CC12M}\xspace}
\newcommand{\coco}{\textsc{MS COCO}\xspace}

\newcommand{\quora}{\textsc{Quora}\xspace}
\newcommand{\fiqa}{\textsc{FiQA}\xspace}
\newcommand{\msmarco}{\textsc{Ms Marco}\xspace}
\newcommand{\beir}{\textsc{BeIR}\xspace}

\newcommand{\nsm}{\textsc{Nsm}\xspace}
\newcommand{\avgclip}{\textsc{AvgClip}\xspace}
\newcommand{\cosinesim}{\textsc{CosineSim}\xspace}
\newcommand{\hessel}{\textsc{HML}\xspace}
\newcommand{\frequency}{\textsc{Freq}\xspace}

\newcommand{\ann}{\textsc{Ann}\xspace}
\newcommand{\auc}{\textsc{Auc}\xspace}

\newtheorem{hypothesis}{Hypothesis}
\newtheorem{definition}{Definition}

\title{Uncovering Visual-Semantic Psycholinguistic Properties from\\the Distributional Structure of Text Embedding Space}

\author{Si Wu\thanks{Both authors contributed equally to this work.} \\
  Northeastern University \\
  \texttt{siwu@ccs.neu.edu} \\\And
  Sebastian Bruch\footnotemark[1] \\
  Northeastern University \\
  \texttt{s.bruch@northeastern.edu} \\}

\begin{document}
\maketitle

\begin{abstract}
Imageability (potential of text to evoke a mental image) and concreteness (perceptibility of text) are two psycholinguistic properties that link visual and semantic spaces. It is little surprise that computational methods that estimate them do so using parallel visual and semantic spaces, such as collections of image-caption pairs or multi-modal models. In this paper, we work on the supposition that text itself in an image-caption dataset offers sufficient signals to accurately estimate these properties. We hypothesize, in particular, that the peakedness of the neighborhood of a word in the semantic embedding space reflects its degree of imageability and concreteness. We then propose an unsupervised, distribution-free measure, which we call \textit{Neighborhood Stability Measure (\nsm)}, that quantifies the sharpness of peaks. Extensive experiments show that \nsm correlates more strongly with ground-truth ratings than existing unsupervised methods, and is a strong predictor of these properties for classification. Our code and data are available on GitHub.\footnote{\url{https://github.com/Artificial-Memory-Lab/imageability}}
\end{abstract}

\section{Introduction}
\label{section:introduction}
The psycholinguistic concept of \emph{imageability}
is a measure of how easily text evokes a clear mental image~\citep{paivio1968concreteness}.
``Beach'' and ``banana'' are examples of highly-\emph{imageable} words.
By contrast, ``gratitude'' and ``criterion'' have low imageability.
Other words fall somewhere in between.

Imageability has been shown to correlate with another psycholinguistic property called \emph{concreteness} \citep{paivio1968concreteness,Richardson1976ImageabilityAC, ellis}. It measures the degree to which a piece of text denotes a perceptible entity~\citep{brysbaert}. Studying and measuring these properties can help us understand how we process text cognitively and further our understanding of perception.

Imageability and concreteness are traditionally measured by surveying
individuals---an expensive and laborious process.
The costs involved have motivated researchers to develop computational methods
to estimate these measures more economically.
As a direct consequence, these methods can help us expand the vocabulary
for which imageability and concreteness ratings are available.

Among the \emph{unsupervised} methods, which is the focus of this work,
one recent study is the work by \citet{wu-smith-2023-composition} which estimates imageability using a text-to-image model. The authors prompt the model with a query and ask it to generate images. They show that the average CLIP scores~\cite{pmlr-v139-radford21a} of generated images correlate with ground-truth imageability ratings.

In an earlier study,~\citet{hessel} propose an unsupervised method to estimate concreteness.
They use a dataset of image-caption pairs and measure how ``clustered'' a word's associated images are.
As they demonstrate, this measure correlates reasonably with ground-truth concreteness ratings.

While these unsupervised methods have lowered the costs of estimating psycholinguistic measures, they each have their own limitations: The method of~\citet{wu-smith-2023-composition} is computationally expensive as it involves sampling, per input text, a number of images from a large, generative model. Sampling too few images can also lead to a high variance in the estimated measurements.

The method of~\citet{hessel} presents its own challenges.
It requires access to a parallel image-caption dataset,
which can be expensive to acquire.\footnote{Many large image-caption collections require researchers to crawl the images independently from the web.}
Additionally, because their method filters images by their
associated words, it is insensitive to word senses and is further susceptible to the \emph{vocabulary mismatch}
problem: Small variations in spellings can affect the measurements.
Indeed, as we will explain later,
this method is able to predict the concreteness of roughly $2\%$ of words for which ground-truth ratings exist.

We set out to address these limitations and present a unified, \emph{unsupervised} approach
for estimating the imageability and concreteness of words.
First, we make the case for \emph{single} modality and aim to quantify
the psycholinguistic properties of interest
using only the captions of an image-caption dataset.
Our hypothesis is that, whether or not words paint a picture
can be determined by the contexts in which they appear;
access to the visual modality is not strictly necessary.

Second, we remedy the vocabulary mismatch problem by operating in the
\emph{semantic embedding} space, similar to what was done by~\citet{wu-smith-2023-composition}.
That is unlike the method by~\citet{hessel}, which uses lexical signals only.

Finally, we hypothesize that, given a sufficiently large collection of
semantic embeddings of text, the structure of the embedding space itself contains clues
on the concreteness and imageability of individual words.
Working directly with the vector collection obviates the need
to sample from the underlying distribution through a generative process.

Importantly, such a vector collection can be constructed once, and reused and recycled indefinitely.
That is in direct contrast to~\citet{wu-smith-2023-composition}, whose method
requires executing multiple rounds of
inference over a large, multi-modal model to quantify the imageability of a single input.

With those objectives in mind, we put forward the following contributions:
\begin{itemize}[leftmargin=*]
    \item We hypothesize that, in a collection of text embeddings,
    the neighborhoods around the embeddings of concrete (imageable) words
    are \emph{structurally different} from those of abstract (non-imageable) words.
    To support our hypothesis, we present an empirical observation that suggests
    the former group is more \emph{stable} than the latter.
    
    \item We present a novel, unsupervised, and distribution-free measure to
    quantify the degree of stability of neighborhoods in an embedding collection, which we call the
    \emph{Neighborhood Stability Measure} (\nsm). As we later show empirically, a piece of text with a more ``stable''
    neighborhood in the embedding space is more imageable or concrete.
    Our method, which is straightforward to implement and fast to execute, draws from the literature
    on approximate nearest neighbor (\ann) search~\cite{Bruch_2024}. 

    \item We conduct a comprehensive experimental evaluation of our proposed measure,
    through which we confirm the validity of our hypothesis above.
    Furthermore, a comparative analysis shows that our algorithm surpasses the performance of
    existing methods by a substantial margin.
\end{itemize}

The remainder of this work is organized as follows: We review the relevant literature in Section~\ref{section:related-work}.
We describe the datasets and embedding models used in this work in Section~\ref{section:datasets-and-embeddings}
to facilitate discussions in subsequent sections.
Section~\ref{section:method} presents our empirical observation and
proposed algorithm. We evaluate our method and compare it with baselines
in Section~\ref{section:evaluation}. We discuss our findings 
in Section~\ref{section:discussion}, before concluding in Section~\ref{section:conclusion}.

\section{Related Work}
\label{section:related-work}

A word's imageability rating is usually obtained by interviewing human subjects, where ratings are on a 7-point Likert scale \citep{toglia1978handbook, gilhooly1980age, Coltheart, wilson}. The process is costly and laborious. Consequently, the resulting database has a limited vocabulary size. The psycholinguistics database where we obtained our ground-truth imageability ratings, MRC Psycholinguistics Database, combines three different large psycholinguistics databases, yet has a vocabulary of $9{,}240$ words~\cite{Coltheart}, of which only $4{,}848$ have imageability ratings.

There have been attempts to expand MRC ratings: \citet{Schock2012ImageabilityEF} added ratings for $3{,}000$ disyllabic words by interviewing human subjects; \citet{liu-etal-2014-automatic-expansion} use synonyms and hyponyms identified in WordNet. The limited success of these attempts presents an opportunity to develop unsupervised computational methods to estimate ratings and expand the vocabulary at a lower cost.

\citet{wu-smith-2023-composition}, for instance, predict the imageability of an input text by prompting a text-to-image model and calculating the average CLIP scores~\cite{pmlr-v139-radford21a} of the generated images. While this method is unsupervised and can estimate the imageability of arbitrary pieces of text, it is computationally expensive as it requires multiple forward passes over a large model per input. Unlike this method, we operate in single modality without the help of a generative model.

\medskip
Separately, the MRC Psycholinguistics Database also contains concreteness ratings of individual words, which have been shown to correlate with imageability ratings~\citep{paivio1968concreteness, Richardson1976ImageabilityAC, ellis}. Later, its vocabulary was expanded to $37{,}058$ words by~\citet{brysbaert} using Amazon Mechanical Turk.

Like imageability, a number of computational methods have been developed to predict concreteness. Among these, \citet{hessel} rely on an image-caption collection to quantify how clustered a word's associated images are, serving as a proxy to its concreteness.

\medskip
The focus of our work is on unsupervised methods, and as such, in our experiments, we focus exclusively on other unsupervised algorithms. However, we would be remiss if we did not note the body of literature on supervised approaches to the task of predicting imageability and concreteness.

\citet{tater-etal-2024-unveiling}, for example, explore the role of \emph{visual} features in determining abstractness versus concreteness, and present a supervised classifier trained on such features. Our method, in contrast, is unsupervised and relies on \emph{textual} features only.

Other works have used lexical and semantic statistics to determine concreteness. \citet{naumann-etal-2018-quantitative}, for instance, are concerned with the distributional differences between abstract and concrete words, but where the distributions are based on part-of-speech tags, and the predictive feature is entropy. We, on the other hand, explore the space of embeddings generated by neural networks, with the predictive feature being the peakedness of distributions in the high-dimensional vector space.

Similarly, the works by \citet{frassinelli-etal-2017-contextual} and \citet{10.3389/frai.2021.796756} all consider distributional properties of text where the distribution is based on co-occurrences (involving part-of-speech tags or other semantic features of individual words). In contrast, the distributions we are concerned with are based on learned representations of text, often trained for determining similarity between pairs.

Other methods include: \citet{Lenci2018-yx} study how Distributional Affective Scores (DAS)---a quantity estimated by a supervised method---differ between concrete and abstract words. The supervised method by \citet{charbonnier-wartena-2019-predicting} trains a regression model over word embeddings and morphological features. Similarly, \citet{kastner} use visual features such as color distribution to estimate imageability. 

\medskip
Finally, the recent study by~\citet{Martinez2024} explores the ability of (private) Large Language Models to predict concreteness. Besides reproducibility limitations due to its dependence on closed-source (and ever-changing) models, this method is also entirely distinct from our unsupervised method that leverages publicly-available datasets and models.

\section{Datasets and Embeddings}
\label{section:datasets-and-embeddings}

Throughout this work, we ground our discussions in specific datasets and embeddings so as to provide illustrative examples and evaluate our proposal. To facilitate that, we introduce these early.

\medskip
\noindent \textbf{Datasets}: As we will explain in subsequent sections, our method estimates the imageability and concreteness
of English words using a (single-modal) text dataset. In particular, we use the text portion of
the datasets listed below:
\begin{itemize}[leftmargin=*]
    \item \textbf{COCO Captions} (\coco): A dataset that contains $1.5$ million captions describing over $330{,}000$ images from the Microsoft Common Objects in Context dataset \citep{lin2015microsoftcococommonobjects}. The image captions were collected using Amazon Mechanical Turk \citep{chen2015microsoftcococaptionsdata}.
    \item \textbf{Conceptual Captions $3$M} (\ccsmall): A dataset that contains $3.3$M filtered image-caption pairs harvested from the web. The description for each image was obtained from the alt-text in the HTML attribute~\citep{sharma-etal-2018-conceptual}. 
    \item \textbf{Conceptual Captions $12$M} (\cclarge): Collected by relaxing the pipeline used to construct \ccsmall: the captions are less precise, but the dataset is $4\times$ larger~\citep{changpinyo2021cc12m}.
\end{itemize}

\medskip
\noindent \textbf{Embedding models}:
Our method operates in the semantic embedding space associated with a (text) dataset;
not in the lexical space. As such, we must transform the datasets described above
into vector collections using text embedding models. We do so using the following
publicly-available models:
\begin{itemize}[leftmargin=*]
    \item \allminilm: This model has been trained on a large number of datasets, exceeding a total of one billion pairs of text.\footnote{Checkpoint available at \url{https://huggingface.co/sentence-transformers/all-MiniLM-L6-v2}}    
    It produces $384$-dimensional embeddings with cosine similarity as distance function. The model is compact in size (with $33$M parameters) and efficient to infer, yet offers competitive quality for semantic retrieval tasks.\footnote{Visit \url{https://sbert.net} for additional details.}
    \item \gtebase and \gtelarge: The \textsc{gte-base-en-v1.5} model and its larger variant \textsc{gte-large-en-v1.5}
    are embedding models with $137$ million and $434$ million parameters respectively~\cite{li2023gte, zhang2024mgte}.\footnote{Checkpoint available at \url{https://huggingface.co/Alibaba-NLP/gte-base-en-v1.5}}\textsuperscript{,}\footnote{Checkpoint available at \url{https://huggingface.co/Alibaba-NLP/gte-large-en-v1.5}}
    The smaller model produces $768$-dimensional embeddings and the larger $1024$-dimensional vectors.
\end{itemize}

The reason we chose these models is two-fold. First, the three models cover a reasonable range of parameter sizes.
Second, the dimensionality of their embedding
space progresses from $384$, to $768$, and finally $1024$. This variety in embedding sizes allows us to examine the
effect of the number of dimensions on the performance of our method.

Finally, we adopt the following naming convention: When dataset $\mathcal{T}$
is embedded with model $\phi$, we denote the resulting vector collection by \vc{$\mathcal{T}$}{$\phi$}.
For example, the collection of \gtebase embeddings of \ccsmall
is denoted by \vc{\ccsmall}{\gtebase}.

\section{Methodology}
\label{section:method}

Consider the imageable word ``beach'' and contrast it with the non-imageable word ``thing.''
Intuitively, the set of contexts in which the former can be used should be more limited than
the contexts in which the latter could appear: ``thing'' can be used in virtually any context.
The same can be said about concrete versus abstract words.
In this section, we develop that intuition into a practical method.

\subsection{Structural clues in the embedding space}
\label{section:method:embedding-space}

Suppose that the intuition above holds, so that the variability of the contexts
a word appears in correlates with its imageability or concreteness.
The first step in formalizing that intuition is to translate it into the ``semantic space'':
A space governed by a distribution from which natural text is sampled.

Translated thus, we arrive at the following statement, which is our hypothesis:
\begin{hypothesis}
\label{hypothesis:main}
The distribution of contexts around an imageable or concrete word
forms a sharper peak in the semantic space.
\end{hypothesis}

That line of reasoning implies that the neighborhoods of words in the semantic space
encode structural clues about the degree of imageability or concreteness.
An imageable word should fall in a more concentrated region of the distribution of text
than a non-imageable word. Indeed, without explicitly noting it,
that is what~\citet{wu-smith-2023-composition} attempt to model, but in a parallel ``image space.''

Unfortunately, we do not have full knowledge of the
structure of the semantic space. Testing the hypothesis above directly is therefore out of the question.
However, there is no shortage of large
text collections made up of rich content, ``sampled'' from a broad region of
the semantic space. Furthermore, with semantic embedding models,
we can transform pieces of text from such collections into vectors.
Such a large collection of embedding vectors can indeed serve as a surrogate to
the semantic space itself, closely capturing its structure.

\begin{figure}[t]
\begin{center}
\centerline{
    \includegraphics[width=0.9\linewidth]{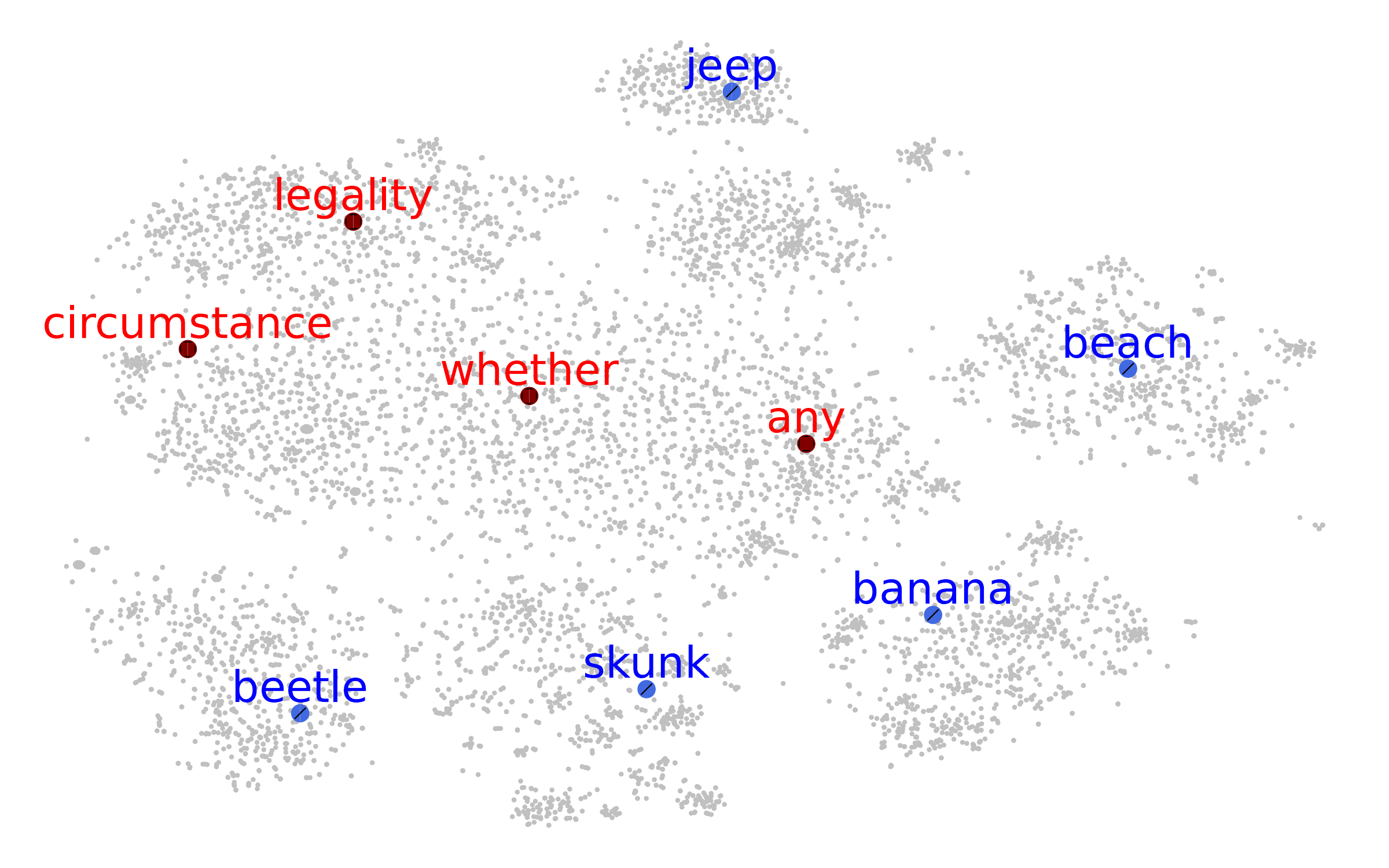}
}
\caption{Visualization of the neighborhoods of a set of imageable words (in blue)
and non-imageable words (in red) in the embedding collection of \vc{\ccsmall}{\allminilm}. We use tSNE~\cite{tsne} for dimensionality reduction. Neighborhoods of non-imageable words have a wider spread and overlap with each other, whereas imageable words have comparatively more separable neighborhoods.}
\label{figure:tsne}
\end{center}
\end{figure}

The method developed and presented next gives us a tool to test Hypothesis~\ref{hypothesis:main}.
Before we proceed, however, we visualize an illustrative example that hints at the
validity of the hypothesis. In Figure~\ref{figure:tsne}, we render the neighborhoods
of select imageable and non-imageable words in the embedding space of \vc{\ccsmall}{\allminilm},
using tSNE~\cite{tsne} for dimensionality reduction.
The neighborhoods of imageable words appear to be more separable
than the neighborhoods of non-imageable words, respectively reflecting the
sharpness and flatness of the peak of the distribution around these words.

\subsection{The Neighborhood Stability Measure}
\label{section:method:nsm}

We design a method to quantify the peakedness of the distribution \emph{around any given neighborhood} in the embedding space
of a text collection. Because we wish for this quantity to serve as a surrogate estimate of the peakedness of the (unknown) distribution that underlies the semantic space, we limit our attention to distribution-free measures. To that end, we draw insights from the literature on approximate nearest neighbors~\cite{Bruch_2024}.

\medskip
\noindent \textbf{Preliminaries}: Consider a semantic embedding space in $\mathbb{R}^d$ equipped with
the similarity measure $\delta: \mathbb{R}^d \times \mathbb{R}^d \rightarrow \mathbb{R}$.
Typically, $\delta(\cdot, \cdot)$ is cosine similarity or inner product.
Denote a \emph{query} point by $q \in \mathbb{R}^d$, representing the embedding of the text whose imageability and concreteness we wish to quantify. Denote further a collection of semantic embeddings, $\mathcal{X} \subset \mathbb{R}^d$, as projections of a large text corpus into the embedding space.

For a fixed $k$, denote by $\mathcal{N}_k(q)$ the subset of $k$ points
from $\mathcal{X}$ that are closest to $q$ according to $\delta$:
\begin{equation}
    \mathcal{N}_k(q) = \argmax^{(k)}_{u \in \mathcal{X}} \delta(q, u),
    \label{equation:ann}
\end{equation}
where the superscript $(k)$ indicates that the argmax
operator returns the $k$ maximizers of its argument.
In practice, we relax $\mathcal{N}_k(q)$ to be the \emph{approximate}
set of nearest neighbors to facilitate scalability.

\medskip
\noindent \textbf{Concepts}: Take a neighborhood $\mathcal{P}$ in some collection $\mathcal{X} \subset \mathbb{R}^d$. We say that $\mathcal{P}$ is $\alpha$-stable, for some $\alpha \in [0, 1]$, if the proportion of points in $\mathcal{P}$ whose nearest neighbor also belongs to $\mathcal{P}$ is $\alpha$.
Formally:

\begin{definition}[$\alpha$-Stable Neighborhood]
\label{definition:alpha-stable-neighborhood}
    A neighborhood $\mathcal{P}$ in $\mathcal{X} \subset \mathbb{R}^d$
    is $\alpha$-stable for $\alpha \in [0, 1]$ calculated as follows:
    \begin{equation}
        \alpha = \frac{1}{\lvert \mathcal{P} \rvert} \Big\lvert \{ u \in \mathcal{P} \;|\; \mathcal{N}_1(u) \in \mathcal{P} \} \Big\rvert,
        \label{equation:alpha-stability}
    \end{equation}
    where $\mathcal{N}_1(\cdot)$ is defined as in Equation~(\ref{equation:ann}) over $\mathcal{X}$,
    and $\lvert \cdot \rvert$ returns the cardinality of its argument.
\end{definition}
This definition is inspired by the concept of ``natural neighbors''~\cite{naturalNeighbor}:
Two points are natural neighbors if they are each others' nearest neighbors.
In effect, we extend that concept to sets of points and quantify it as a continuous measure.

\begin{figure}[t]
\begin{center}
\centerline{
  \includegraphics[width=0.7\linewidth]{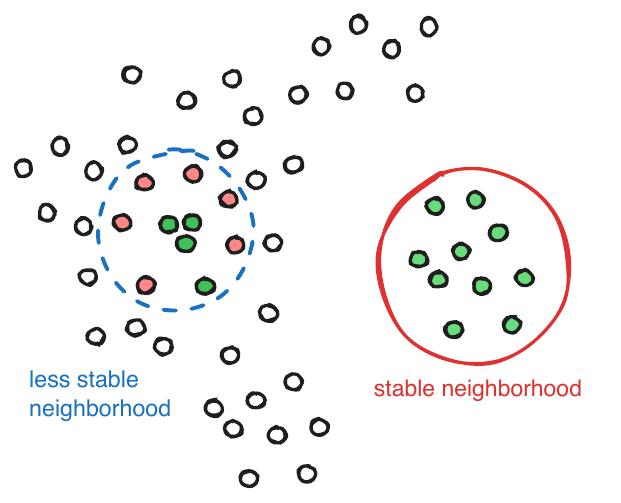}
}
\caption{Illustration of a stable neighborhood (encircled by solid red boundary) versus a less stable neighborhood (inside the dashed blue region). The nearest neighbor of every point in the former belongs to the same neighborhood, whereas only $4$ out of $10$ points in the latter have that property. If these points are sampled from an unknown distribution, the \emph{stability} of these neighborhoods reflects the sharpness or flatness of the regions around them in the underlying distribution.}
\label{figure:neighborhood-stability}
\end{center}
\end{figure}

Let us take a moment and understand what $\alpha$-stability says about a neighborhood visually. In Figure~\ref{figure:neighborhood-stability},
we illustrate two neighborhoods. One with perfect stability (i.e., $\alpha=1$)
and another with stability factor $\alpha=0.4$. The former is separable from the rest of the collection, while the latter is surrounded by other points. Assuming that these points are drawn from a distribution, the separability or inseparability of a neighborhood is telling of the sharpness or flatness of the corresponding regions in the distribution.

\medskip
Taking $k$ as a hyperparameter and assuming access to an approximate nearest neighbor (\ann) search algorithm,
we can measure the $\alpha$-stability of the neighborhood in a collection $\mathcal{X}$ around
any arbitrary query point $q$. We call this quantity $q$'s Neighborhood Stability Measure with radius $k$,
defined formally as follows:

\begin{definition}[Neighborhood Stability Measure for point $q$ and radius $k$]
\label{definition:nsm}
Suppose $\mathcal{N}_k(q)$ is an $\alpha$-stable neighborhood.
Then the Neighborhood Stability Measure (\nsm) of $q$ given radius $k$ is $\alpha$.
\end{definition}

\medskip
\noindent \textbf{Algorithm}: We use the \nsm of the query point $q$ with radius $k$ as an estimate of the imageability and concreteness of the text represented by $q$. We leave $k$ as a hyperparameter to be tuned using
a validation set of queries given a text collection.
Note that, an \nsm closer to $1$ indicates a higher imageability and concreteness, as per Hypothesis~\ref{hypothesis:main}.

The algorithm to compute the \nsm of a query is described in Algorithm~\ref{algorithm:nsm}. The complexity of a na\"ive implementation of that is dominated by $\mathcal{O}(k T)$ where $T$ is the complexity of the \ann search subroutine. In a practical implementation, one can populate a table mapping each point in the collection to its (approximate) nearest neighbor upfront; the call to the \ann function on Line~\ref{algorithm:nsm:inner-ann} of Algorithm~\ref{algorithm:nsm} can then be replaced with a table lookup, reducing the complexity of the algorithm to $\mathcal{O}(T)$. We note that, state-of-the-art \ann algorithms are highly efficient and can easily scale to collections of hundreds of millions of points~\cite{Bruch_2024}.

\begin{algorithm}[t]
\caption{Neighborhood Stability Measure}
\label{algorithm:nsm}
\textbf{Input}: Query $q \in \mathbb{R}^d$; embedding collection $\mathcal{X} \in \mathbb{R}^d$; radius, $k$; subroutine $\ann_\mathcal{X}(q, k)$ that returns the $k$ approximate nearest neighbors of $q$ from $\mathcal{X}$.

\textbf{Output}: \nsm of $q$ with radius $k$.
\begin{algorithmic}[1]
    \State $\mathcal{N}_k(q) \leftarrow \ann_\mathcal{X}(q, k)$
    \State $\alpha \leftarrow 0$
    \For{$u \in \mathcal{N}_k(q)$}
        \If{$\ann_\mathcal{X}(u, 1) \in \mathcal{N}_k(q)$} \label{algorithm:nsm:inner-ann}
            \State $\alpha \leftarrow \alpha + \frac{1}{k}$
        \EndIf
    \EndFor
    \Return $\alpha$
\end{algorithmic}
\end{algorithm}

\section{Empirical Evaluation}
\label{section:evaluation}

We evaluate our proposal in this section. We explain our setup and present our results.

\subsection{Setup}
\label{section:evaluation:setup}
We have already introduced the datasets and embedding models used in this work in Section~\ref{section:datasets-and-embeddings}.
Here, we explain the dataset of ground-truth imageability and concreteness ratings;
our evaluation protocol; and, the methods that will be put to the test.

\medskip
\noindent \textbf{Ground-truth ratings}: We use imageability ground-truth ratings for $4{,}848$ words from the MRC Psycholinguistics Database \citep{Coltheart} collected by interviewing human subjects. The concreteness ground-truth ratings by \citet{brysbaert} were crowd-sourced with Amazon Mechanical Turk, with a vocabulary size of $37{,}058$.

\medskip
\noindent \textbf{Evaluation protocol}:
We randomly split each ratings dataset into non-overlapping validation ($20\%$)
and test ($80\%$) subsets. If a method requires hyperparameter tuning,
we do so on the validation split. We evaluate all methods on the test set only.
The measures reported in this work are the average of $10$ trials, where each trial
uses its own seed for the pseudo-random generator that produces the splits.

We evaluate all methods on the imageability axis as follows:
For every word from the imageability ratings dataset, we compute its
score according to each method. We then measure the Spearman's correlation
coefficient between the ratings and predictions.
Evaluation on the concreteness axis follows the exact same logic, except that
the set of words come from the concreteness ratings dataset.

\medskip
\noindent \textbf{Methods}:
We compare our proposed measure with a number of existing methods from the literature.
The complete list of methods is as follows:
\begin{itemize}[leftmargin=*]
    \item $\frequency_\mathcal{T}$: This method works with the lexicon only.
    If we are interested in estimating the imageability or concreteness
    of the word $w$, we simply count the number of times $w$ appears in the text collection $\mathcal{T}$
    and use that as the prediction. This na\"ive baseline mirrors what was presented in
    prior work~\cite{wu-smith-2023-composition,hessel}. Here, $\mathcal{T}$ is the collection of captions from \coco, \ccsmall, and \cclarge.
    \item $\hessel_\mathcal{D}$: This denotes the measure developed by~\citet{hessel} to predict the \emph{concreteness} of words; the acronym represents the initial letters of the authors' surname. The subscript denotes an image-caption dataset,
    over which the measure is calculated. In our experiments, we use \coco as $\mathcal{D}$ as per the original work.
    The concreteness score of a word is computed by counting how often the nearest neighbors of an associated image are also associated with the same word. We use the implementation recommended by the author to conduct our experiments.\footnote{For details, visit \url{https://jmhessel.com/projects/concreteness/concreteness.html}.}
    
    \item \avgclip and \cosinesim: Measures proposed by~\citet{wu-smith-2023-composition}
    to predict the \emph{imageability} of words and phrases.
    To estimate that quantity for word $w$, we use the code provided by the authors to present $w$ to a text-to-image model, DALL•E mini,
    and ask it to generate $16$ images.\footnote{\url{https://github.com/swsiwu/composition_and_deformance}} \cosinesim computes the average pairwise cosine similarity
    between the embeddings of the generated images, while \avgclip reports the mean CLIP scores~\cite{pmlr-v139-radford21a}.

    \item $\nsm_\mathcal{X}$: This is our proposal. When we measure
    \nsm on the vector collection $\mathcal{X}$, we denote it by $\nsm_\mathcal{X}$,
    where $\mathcal{X}$ is the embeddings of \coco, \ccsmall, or \cclarge.
    We use the Faiss library~\cite{douze2024faisslibrary} to perform \ann search using an Inverted File (IVF) index, where the number of clusters is set to $8 \sqrt{\lvert \mathcal{X} \rvert}$
    where $\lvert \cdot \rvert$ is the size of its argument.
    During search, the parameter \texttt{nprobe} is set to $128$; this results in a near exact nearest neighbor search
    over the datasets in our experiments. Finally, we tune the radius $k$ of Algorithm~\ref{algorithm:nsm} on the validation set, by sweeping the interval $[64, 4096]$ in increments of $64$---we examine the effect of $k$ in Appendix~\ref{appendix:radius}.
\end{itemize}

\subsection{Results}
\label{section:evaluation:results}

\begin{table}[t]
\small
\begin{center}
\begin{sc}
\begin{tabular}{l|cc}
\toprule
Measure & Imag. & Conc. \\
\midrule
$\frequency_\coco$ & $0.26$ & $0.25$ \\
$\frequency_\ccsmall$ & $0.23$ & $0.30$ \\
$\frequency_\cclarge$ & $0.34$ & $0.35$ \\
$\hessel_\coco$ & $0.49$ & $0.45$ \\  
\cosinesim & $0.45$ & $0.40$ \\
\avgclip & $0.56$ & $0.45$ \\ 
\midrule
$\nsm_{\vc{\coco}{\allminilm}}$ & $0.57$ & $0.53$ \\
$\nsm_{\vc{\ccsmall}{\allminilm}}$ & $0.61$ & $0.55$ \\
$\nsm_{\vc{\cclarge}{\allminilm}}$ & $\mathbf{0.66}$ & $\mathbf{0.58}$ \\
\midrule
$\nsm_{\vc{\coco}{\gtebase}}$ & $0.54$ & $0.54$ \\
$\nsm_{\vc{\ccsmall}{\gtebase}}$ & $0.57$ & $0.58$ \\
$\nsm_{\vc{\cclarge}{\gtebase}}$ & $0.58$ & $0.58$ \\
\midrule
$\nsm_{\vc{\coco}{\gtelarge}}$ & $0.45$ & $0.44$ \\
$\nsm_{\vc{\ccsmall}{\gtelarge}}$ & $0.56$ & $0.51$ \\
$\nsm_{\vc{\cclarge}{\gtelarge}}$ & $0.56$ & $0.53$ \\
\bottomrule
\end{tabular}
\end{sc}
\end{center}
\caption{Spearman correlation coefficient between ratings and predictions for imageability (Imag.) and concreteness (Conc.). The Neighborhood Stability Measure (\nsm) is ours, with the subscript denoting the collection it was measured on. Baselines in the first block include~\citet{hessel} ({\small \hessel});~\citet{wu-smith-2023-composition} ({\small \cosinesim and \avgclip}); and, word frequency ({\small\frequency}).}
\label{table:correlations}
\end{table}

We present our main results in Table~\ref{table:correlations}, where we report the correlation between the ground-truth imageability and concreteness ratings with the predictions of each method. We note that, even though \avgclip and \cosinesim are designed to measure imageability, we measure their correlation on the concreteness dataset as well for completeness. Similarly, we measure the performance of \hessel on the imageability dataset, despite the fact that it was intended for estimating concreteness.

\medskip
We find that \frequency yields a stronger-than-expected correlation with both imageability and concreteness. This was not the case in the experiments of~\citet{wu-smith-2023-composition}, who report no correlation. That is because, in their experiments,~\citet{wu-smith-2023-composition} use word frequency from the Brown Corpus \citep{francis79browncorpus}, which is a general corpus of natural texts. Here, however, we measure the correlation between the frequency of words in a collection of captions, resulting in a higher correlation and a stronger baseline. We note that, unsurprisingly, larger collections lead to stronger correlation coefficients, with one exception ($\frequency_\ccsmall$ versus $\frequency_\coco$).

\begin{table}[t]
\small
\begin{center}
\begin{sc}
\begin{tabular}{l|cc}
\toprule
Measure & Imag. & Conc. \\
\midrule
$\frequency_\coco$ & $67.6\%$ & $28.7\%$ \\
$\frequency_\ccsmall$ & $84.5\%$ & $47.6\%$ \\
$\frequency_\cclarge$ & $98.0\%$ & $71.6\%$ \\
$\hessel_\coco$ & $12.9\%$ & $2.7\%$ \\
\bottomrule
\end{tabular}
\end{sc}
\end{center}
\caption{Percentage of words with ground-truth ratings for which each method is able to predict imageability and concreteness. \cosinesim, \avgclip, and \nsm offer a coverage of $100\%$ by design.}
\label{table:coverage}
\end{table}

\medskip
$\hessel$ appears to perform better than $\frequency$; However, the method is able to predict the imageability of only about $13\%$ of words in the ratings dataset. Similarly, it only predicts concreteness scores for $2.7\%$ of total words. That is due to the vocabulary mismatch problem: the vast majority of words in the ratings dataset do not appear in the captions. As we explained, that is a major limitation of the method of~\citet{hessel}. Table~\ref{table:coverage} reports the coverage for all methods.

Finally, \avgclip performs better on the imageability dataset, confirming the findings of~\citet{wu-smith-2023-composition}. Its performance on the concreteness prediction task is also relatively strong. This method, however, is far more computationally expensive than the other baselines. Using two GeForce RTX 3080 Ti GPUs, it took approximately 120 hours to compute all concreteness scores.

\medskip
We now turn to our method. Every configuration of \nsm outperforms the strongest baseline on both imageability and concreteness tasks, with two exceptions. That goes some way to validate Hypothesis~\ref{hypothesis:main}, lending credit to the assertion that the embedding space contains structural clues on imageability and concreteness.

In addition to the overall performance, we highlight two important trends. First, the size of the embedding collection matters: regardless of what embedding model is used, larger collections often outperform smaller collections. That is easy to explain: A larger collection represents more granular samples from the underlying distribution, painting a more accurate picture of its structure.

Second, interestingly, the dimensionality of the space plays an important role: predictions correlate more strongly as the dimensionality of the space becomes smaller. The reason for this phenomenon is simple and has to do with the curse of dimensionality: As dimensionality increases, the pairwise distances between points concentrate more heavily around a constant~\cite{Bruch_2024}, distorting the picture painted by sample points (i.e., the embedding collection) of the sharpness and flatness of the peaks of the local distribution. As such, an ever larger collection is needed to accurately discover distributional structures of the embedding space.

We conclude our analysis of the results with one last point: as with \frequency, it matters to our method that the text collections from which \nsm is calculated are not general natural texts, and instead contain text fragments that describe visual phenomena. That is for good reason: imageability and concreteness are visual-semantic properties. It is therefore reasonable that estimating these properties would require the ``visual regions'' of the semantic space. 

\begin{table}[t]
\begin{center}
\begin{sc}
\small
\begin{tabular}{l|cc}
\toprule
Measure & Imag. & Conc. \\
\midrule
$\nsm_{\vc{\cclarge}{\allminilm}}$ & $\mathbf{0.66}$ & $\mathbf{0.58}$ \\
\midrule
$\nsm_{\vc{\quora}{\allminilm}}$ & $0.40$ & $0.31$ \\
$\nsm_{\vc{\fiqa}{\allminilm}}$ & $0.29$ & $0.33$ \\
$\nsm_{\vc{\msmarco}{\allminilm}}$ & $0.48$ & $0.39$ \\
\bottomrule
\end{tabular}
\end{sc}
\end{center}
\caption{Spearman correlation between ground-truth
imageability (Imag.) and concreteness (Conc.) ratings and \nsm
using non-caption text datasets. \quora (question answering, $523$K documents)
and \fiqa (financial, $57$K) are from the \beir collection~\cite{thakur2021beir},
and \msmarco~\cite{nguyen2016msmarco} is a dataset of about $9$ million passages designed for retrieval tasks.
}
\label{table:general-purpose-collection}
\end{table}

For completeness, we verify the above by repeating our experiments on general-purpose text collections. As Table~\ref{table:general-purpose-collection} shows, \nsm's correlation degrades when the text collection is not primarily made up of image captions. This agrees with the findings of \citet{wu-smith-2023-composition} where the highest correlation with imageability ratings was achieved by image captions, not news sentences or poems.

\subsection{Analysis}
\label{section:evaluation:analysis}

We observed that \nsm correlates with ground-truth ratings more strongly than existing unsupervised methods, often with a substantial margin. However, the correlation coefficient itself is less than ideal, implying perhaps that today's unsupervised methods cannot fully replace the manual human judgment process. The reality, however, is more nuanced, as our analysis in this section suggests.

Consider a classification task where we place a word into imageable (concrete) or non-imageable (abstract) categories. As in prior studies~\cite{10.3389/frai.2021.796756,tater-etal-2024-unveiling}, we define ground-truth labels by thresholding the ratings. For instance, words whose imageability ratings are greater than $\theta$ would be considered imageable, and those below it non-imageable. \nsm scores can therefore be interpreted as class membership probabilities so that for any given $\theta$, we can calculate the Area Under the Receiver Operating Characteristic Curve (\auc).

\begin{figure}[t]
\begin{center}
\centerline{
    \subfloat[Imageability]{
        \includegraphics[width=0.47\linewidth]{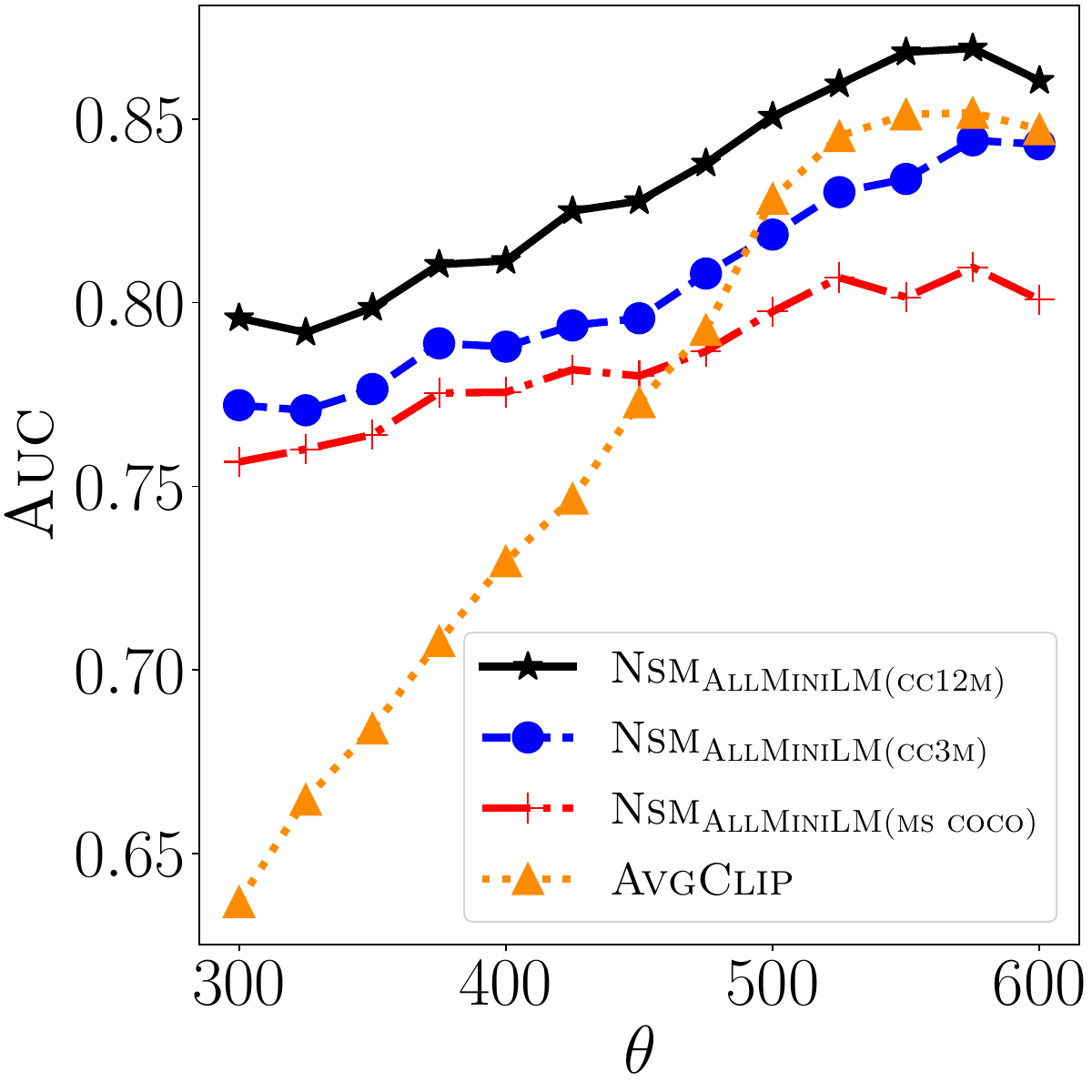}
    }
    \subfloat[Concreteness]{
        \includegraphics[width=0.47\linewidth]{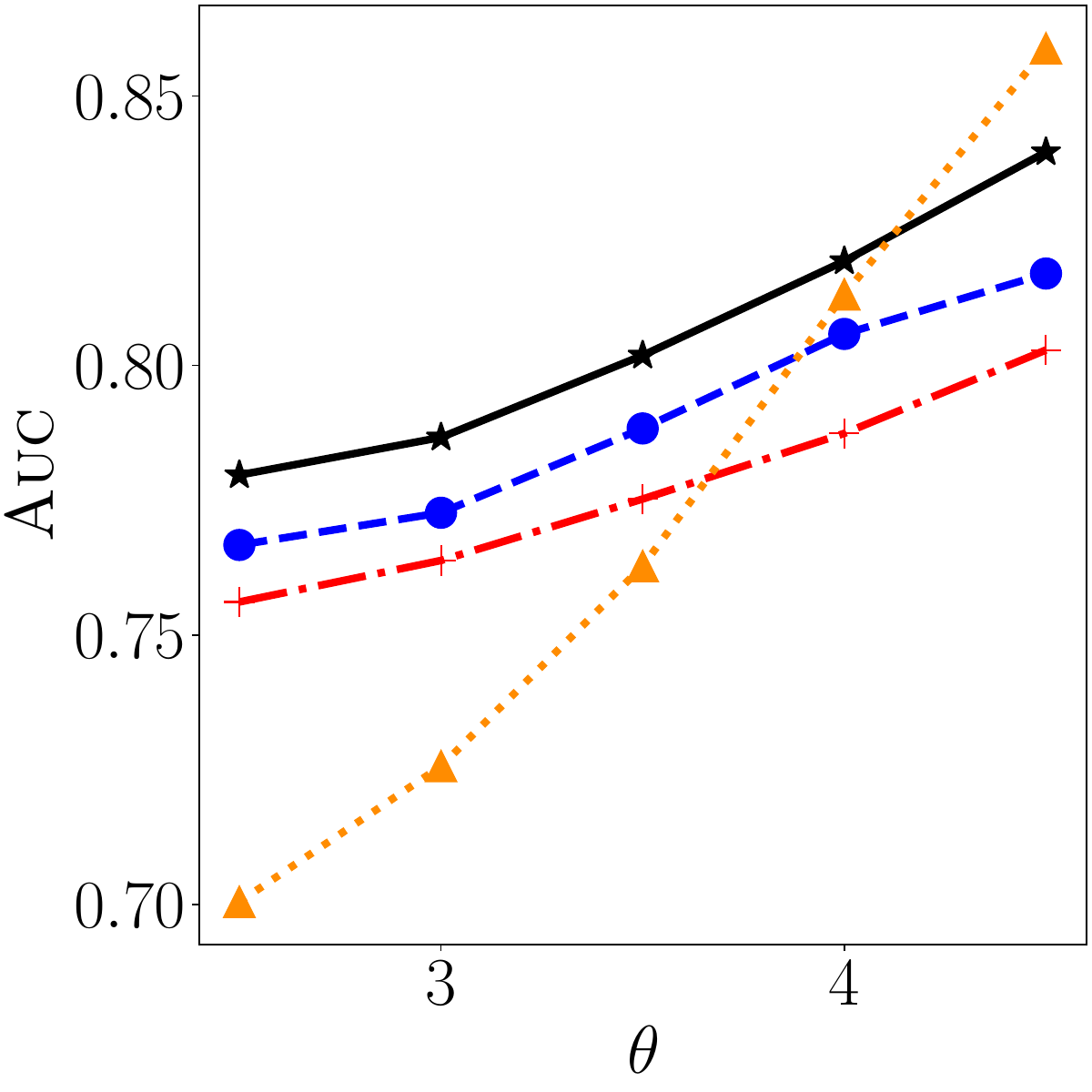}
    }
}
\caption{Performance in classification tasks. The threshold $\theta$ determines class labels: A word whose ground-truth rating is less than $\theta$ is non-imageable (abstract). For each $\theta$, we evaluate the effectiveness of select methods as class predictors, quantified using \auc.}
\label{figure:auc}
\end{center}
\end{figure}

Figure~\ref{figure:auc} visualizes \auc for select methods and different values of $\theta$. The plots suggest that \nsm serves as a strong predictor of imageability and concreteness in classification tasks. Furthermore, its predictive power is relatively robust to the choice of $\theta$: Even though the correlation coefficients of \avgclip and $\nsm_{\vc{\coco}{\allminilm}}$ are similar for imageability, the former's predictions are far more sensitive to how we derive class labels.

Let us unpack the predictions by $\nsm_{\vc{\cclarge}{\allminilm}}$ further and study the reliability of the \nsm scores. We choose a margin, $\omega \in (0, 0.5]$, and exclude words whose \nsm score is in the interval $(\omega, 1 - \omega)$. We then compute \auc as before. In this way, we seek to understand if the extreme values offer more predictive power and, thus, are more meaningful.

\begin{figure}[t]
\begin{center}
\centerline{
    \subfloat[Imageability]{
        \includegraphics[width=0.47\linewidth]{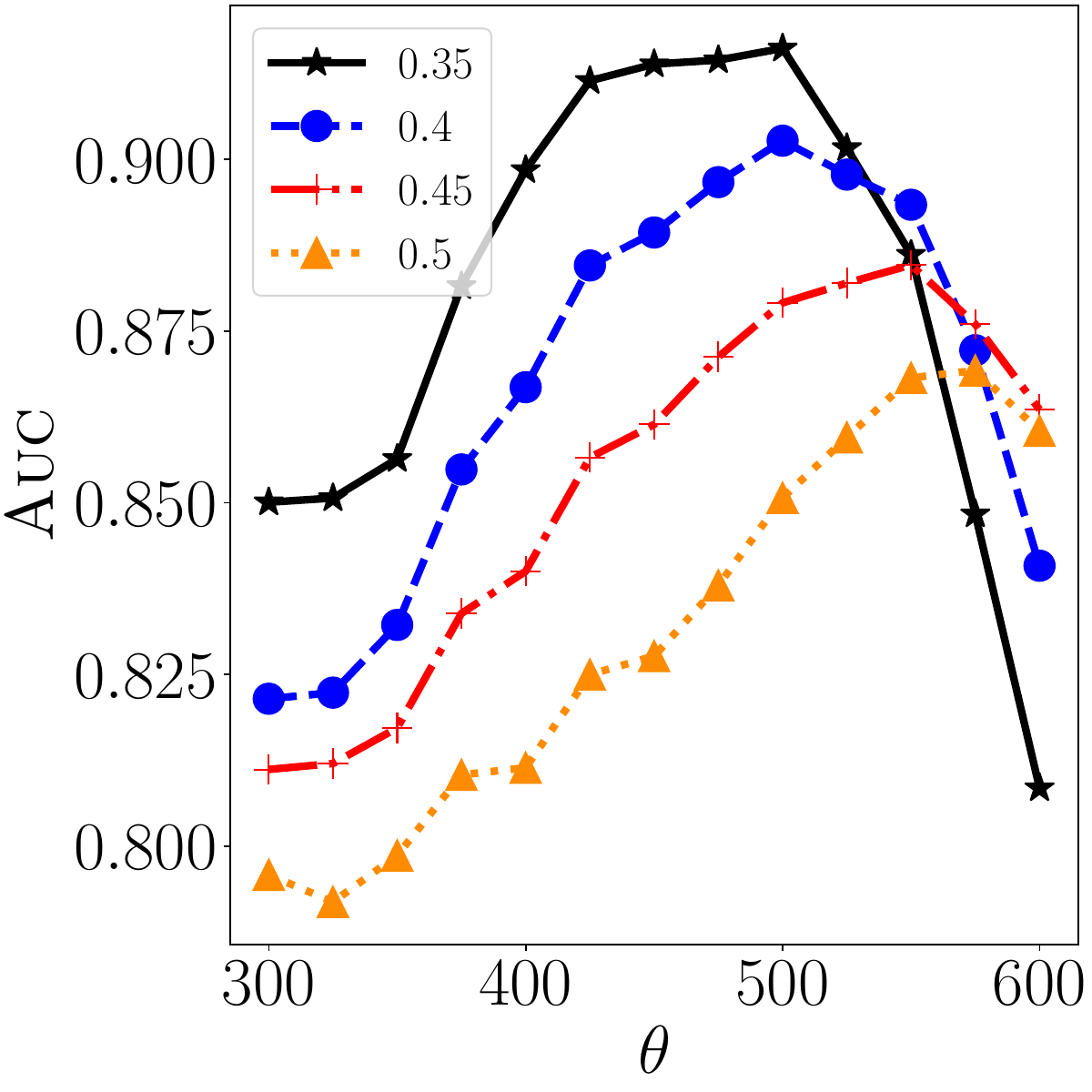}
    }
    \subfloat[Concreteness]{
        \includegraphics[width=0.47\linewidth]{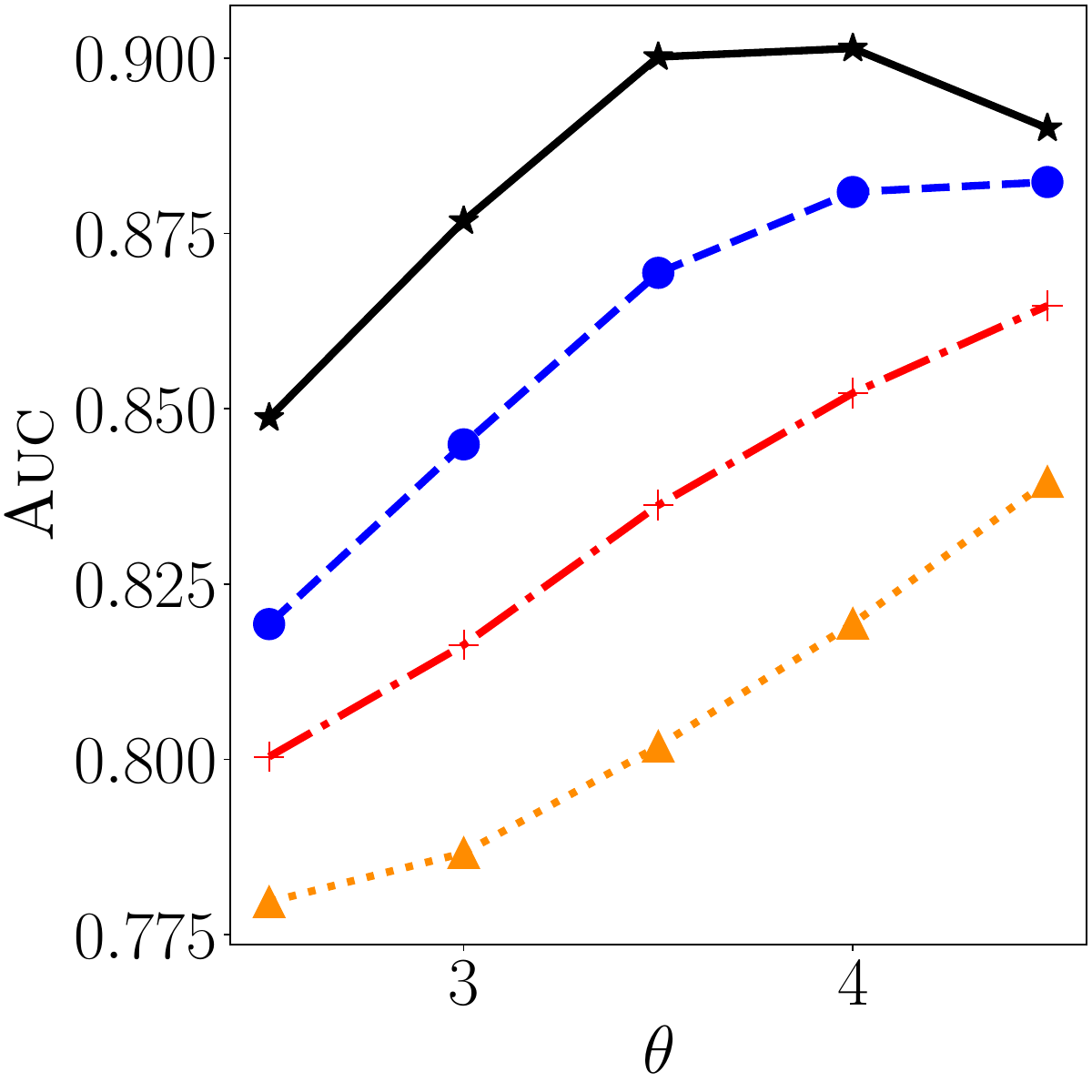}
    }
}
\caption{Classification by $\nsm_{\vc{\cclarge}{\allminilm}}$. Words whose \nsm scores are in $[0, \omega]$ are negative (i.e., non-imageable or abstract), and $[1 - \omega, 1]$ positive, repeated for $\omega \in \{0.35, 0.4, 0.45, 0.5\}$. There is more confidence in predictions if the \nsm score is very small or very large.}
\label{figure:margin-auc}
\end{center}
\end{figure}

Figure~\ref{figure:margin-auc} shows this analysis visually for select margins---margins smaller than $0.35$ result in too few samples. Indeed, as the margin becomes smaller, classification accuracy improves. That trend suggests that, if the \nsm score for a word is very small (large), the odds of it being non-imageable (imageable) are stronger.

\section{Discussion}
\label{section:discussion}

We have shown that Algorithm~\ref{algorithm:nsm} serves as a tool for testing Hypothesis~\ref{hypothesis:main}, and that its predictions correlate with and are predictive of our visual-semantic psycholinguistic properties of interest. Our measure, \nsm, also operates considerably more efficiently than existing unsupervised methods.

Let us expand on efficiency. \nsm is \emph{data efficient} as it requires only the text portion of an image-caption dataset, while \hessel requires a corpus of image-caption pairs. \nsm is \emph{computationally efficient} as it operates on a (reusable) collection of text embeddings, performing only \ann search over it. In contrast, the \avgclip or \cosinesim methods of~\citet{wu-smith-2023-composition} execute multiple forward passes over an expensive text-to-image model to predict the imageability of a single input text.

Moreover, \nsm is \emph{space efficient} as it needs 1) a table mapping the identifier of each point in the collection $\mathcal{X}$ to the identifier of its nearest neighbor, which can be stored in an array of length $\lvert \mathcal{X} \rvert$; and 2) an \ann index, which is typically a fraction of the size of the collection. For example, the size of the \ann index for \vc{\cclarge}{\allminilm} is $938$MB while the embedding collection itself is $9.5$GB in size. \vc{\cclarge}{\gtelarge} is $25.4$GB in size, while its \ann index is $1$GB.

It is worth noting that we use contextualized embeddings of a token, where a token can have diverse contexts and part-of-speech (POS) tags in an image-caption dataset. In contrast, context and POS tags were not indicated when collecting ratings from surveyed humans. Nevertheless our method estimates ratings effectively, suggesting perhaps that a large image-caption dataset captures the everyday context of a word to some extent. Though we do not offer an answer in this work, that raises an interesting psycholinguistic question: do humans consider diverse contexts when rating imageability?

Lastly, we note that our method works in the embedding space and as such is agnostic to the characteristics of the input prior to embedding. That entails that Algorithm~\ref{algorithm:nsm} can in theory be applied to a sentence, after it is projected into the embedding space. We are, however, unable to demonstrate this extension for two reasons. First, to the best of our knowledge, no dataset exists that provides the imageability rating of sentences. Second, how a sentence is perceived is more complex as it involves compositionality. We believe an extensive qualitative study involving surveys for sentence-level imageability makes for an exciting future direction.

\section{Conclusion}
\label{section:conclusion}

We hypothesized that text collections provide sufficient clues on visual-semantic psycholinguistic properties of words. In particular, we claimed that the range of contexts in which a word appears correlates with imageability and concreteness. We then formalized that claim in distributional terms in Hypothesis~\ref{hypothesis:main}, stating that the structure of the embedding space of a sufficiently large text collection encodes imageability and concreteness.

We subsequently developed a method to quantify the ``sharpness'' of peaks in the distribution---the Neighborhood Stability Measure. A perfectly-stable neighborhood, according to Definition~\ref{definition:alpha-stable-neighborhood}, is one where the nearest neighbor of every point belongs to the same neighborhood. More stable neighborhoods correspond with sharper peaks. Empirically, \nsm correlates more strongly with imageability and concreteness ratings than all other unsupervised baselines. Additionally, \nsm is a strong predictor of class membership in classification tasks. Finally, \nsm is comparatively efficient, and addresses the shortcomings of prior methods.

Even though we have demonstrated that single-modality proves adequate, we ask whether bringing visual features into a joint embedding space can enrich the distributional structure, and lead to better predictions. We have also left unanswered the question of \emph{why} \nsm performs better on image-caption datasets, and \emph{whether} its performance can be improved on general natural text collections. Finally, while we claim that our method is applicable to text fragments (as opposed to single words), we have not tested that claim. We leave an exploration of these directions as future work.

\newpage
\section*{Limitations}
1. Our experiments were done for English words only. We cannot definitely conclude whether similar observations can be made when applying our method to languages that are morphologically different. \\
2. We have shown in our experiments that our method relies on text collections that describe visual phenomena such as image captions. Such datasets are not as prevalent as general-purpose natural text collections, especially in languages other than English. As such, extending our findings to other languages can face practical challenges.

\section*{Potential risks}
To obtain \avgclip and \cosinesim, we replicated the experiments by \citet{wu-smith-2023-composition} which utilize a text-to-image model. As \citet{wu-smith-2023-composition} claimed, DALL•E mini can potentially generate suggestive images even when given a text that does not contain suggestive content. While we did not notice any images generated that were suggestive, one should be cautious about the output of a text-to-image model, especially if they intend to release the images. 

\section*{Acknowledgments}
We extend our sincerest gratitude to all the anonymous reviewers and meta-reviewers who meticulously reviewed our work and gave us constructive feedback. Their balanced review raised valid concerns but recognized our proposal's strength. Their suggestions led to much needed clarification in the prose and several notable improvements to our evaluation design. We appreciate their time and the effort they put into the review process.

\bibliography{main}

\appendix

\section{Effect of radius}

The Neighborhood Stability Measure (\nsm) has a single hyperparameter, $k$, which represents the size of the neighborhood for which we calculate the stability measure. In our main experiments, we used a validation set to tune $k$ by sweeping the interval $[64, 4096]$ in increments of $64$. In this supplementary section, we examine the effect $k$ has on the correlation between \nsm and imageability, as well as \nsm and concreteness.

\begin{figure}[t]
\begin{center}
\centerline{
    \subfloat[Imageability]{
        \includegraphics[width=0.47\linewidth]{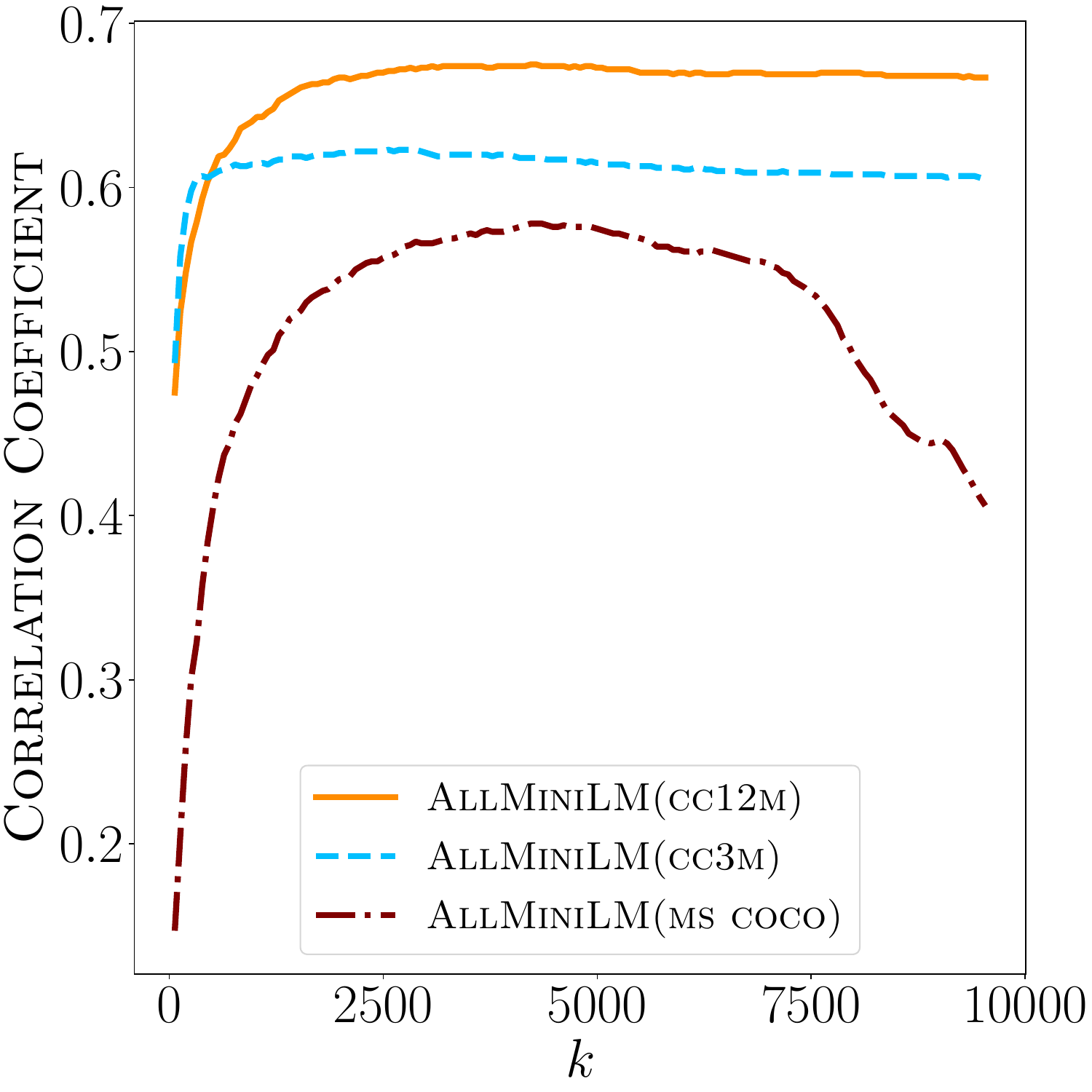}
    }
    \subfloat[Concreteness]{
        \includegraphics[width=0.47\linewidth]{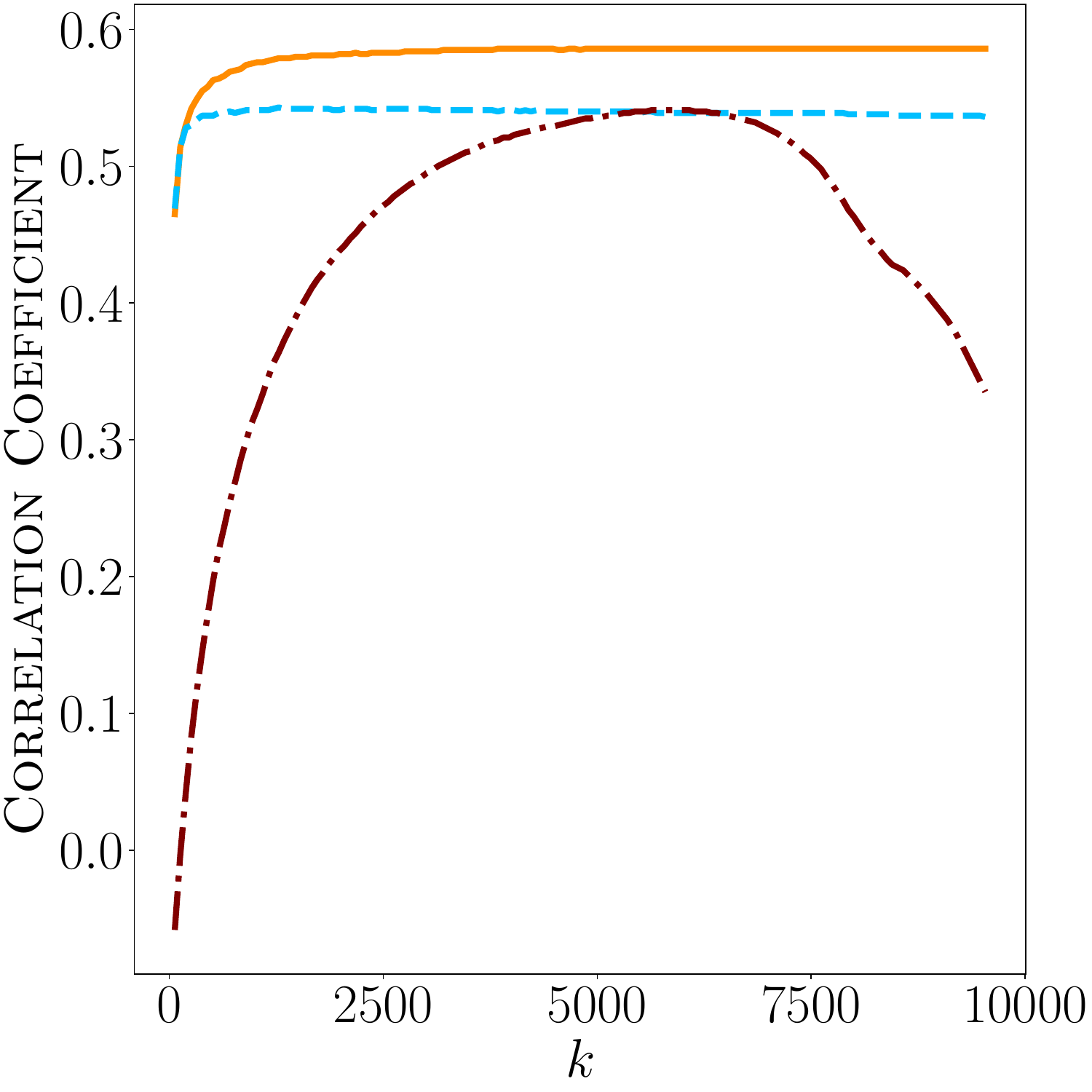}
    }
}
\caption{Effect of the \nsm radius, $k$, on the correlation with imageability and concreteness, for select embedding collections. A value between $2^{10}$ and $2^{12}$ yields peak performance for reasonably large collections.}
\label{figure:radius}
\end{center}
\end{figure}

Figure~\ref{figure:radius} illustrates the Spearman's correlation coefficient as a function of $k$. In other words, for each value of $k$, we compute the \nsm scores for all words in the imageability and concreteness ratings datasets, and compute the correlation of the scores with ground-truth ratings. Note that, while Figure~\ref{figure:radius} visualizes this result for \allminilm embeddings, other embedding collections exhibit a similar trend.

As the plots show, a small value of $k$ leads to poor correlation. The reason behind this is that, \nsm takes a value in $\{ 0, \frac{1}{k}, \frac{2}{k}, \ldots, 1 \}$. When $k$ is too small, the score set becomes too coarse. On the other hand, the imageability and concreteness ratings are fine-grained. This discrepancy between the granularity of \nsm score set and the ratings leads to poor correlation, as \nsm is unable to distinguish between ratings that are close to each other.

As $k$ becomes larger, correlation stabilizes. That trend continues until $k$ is too large relative to the size of the collection. Interestingly, independent of the size of the dataset, $k \in [2^{10}, 2^{12}]$ appears to yield peak correlation. A $k$ in that range can thus serve as a reasonable default value where a validation set is unavailable.

\label{appendix:radius}

\end{document}